%% file: main.tex
\documentclass[acmtog]{acmart}
\acmSubmissionID{66}

\usepackage{booktabs} % For formal tables

\input{preamble}

\usepackage[ruled]{algorithm2e} % For algorithms

\SetAlFnt{\small}
\SetAlCapFnt{\small}
\SetAlCapNameFnt{\small}
\SetAlCapHSkip{0pt}

\AtBeginDocument{%
  }

\copyrightyear{2025}
\acmYear{2025}
\setcopyright{acmlicensed}\acmConference[SIGGRAPH Conference Papers '25]{Special Interest Group on Computer Graphics and Interactive Techniques Conference Conference Papers }{August 10--14, 2025}{Vancouver, BC, Canada}
\acmBooktitle{Special Interest Group on Computer Graphics and Interactive Techniques Conference Conference Papers (SIGGRAPH Conference Papers '25), August 10--14, 2025, Vancouver, BC, Canada}
\acmDOI{10.1145/3721238.3730655}
\acmISBN{979-8-4007-1540-2/2025/08}

\begin{document}

% Title portion
\title{Scene-Level Appearance Transfer with Semantic Correspondences}
\author{Liyuan Zhu}
% \authornote{Equal contribution}
\email{liyzhu@stanford.edu}
\affiliation{%
 \institution{Stanford University}
 \city{Stanford}
 \country{USA}}

 \author{Shengqu Cai}
\authornote{Equal contribution}
\email{shengqu@stanford.edu}
\affiliation{%
 \institution{Stanford University}
 \city{Stanford}
 \country{USA}}

 \author{Shengyu Huang}
\authornotemark[1]
\email{shengyuh@nvidia.com}
\affiliation{%
 \institution{NVIDIA Research}
 \city{Zurich}
 \country{Switzerland}}

 \author{Gordon Wetzstein}
\email{gordon.wetzstein@stanford.edu}
\affiliation{%
 \institution{Stanford University}
 \city{Stanford}
 \country{USA}}

 \author{Naji Khosravan}
\email{najik@zillow.com}
\affiliation{%
 \institution{Zillow Group}
 \city{Seattle}
 \country{USA}}

 \author{Iro Armeni}
\email{iarmeni@stanford.edu}
\affiliation{%
 \institution{Stanford University}
 \city{Stanford}
 \country{USA}}
% DO NOT ENTER AUTHOR INFORMATION FOR ANONYMOUS TECHNICAL PAPER SUBMISSIONS TO SIGGRAPH 2019!
\input{figures/tex/teaser}
\input{sections/1_abstract}




\maketitle

\input{sections/2_introduction}
\input{sections/3_related_work}

\input{sections/4_method}

\input{sections/5_experiments}

\input{sections/6_conclusion}
\begin{acks}
We thank Bingxin Ke, Qinxin Yan, Yuru Jia, and Emily Steiner for the fruitful discussion, and Jianhao Zheng for the help in making the video. 
\end{acks}
\bibliographystyle{ACM-Reference-Format}
\bibliography{egbib}

\clearpage
\input{figures/tex/refinement}
\input{figures/tex/2d_transfer_extra}
\input{figures/tex/style_lifting}
\input{figures/tex/extra_end2end}

% commenting out appendix for compiling speed

% \input{sections/7_appendix}

\end{document}

%% file: preamble.tex
\usepackage{multirow} % used by tables
\usepackage{makecell} % used by tables
\usepackage{bm}
\usepackage{subcaption} % subtable

\def\eg{\emph{e.g.}\xspace} 
\def\ie{\emph{i.e.}\xspace} 

\def\wrt{\emph{w.r.t.}\xspace}

\newcommand{\ours}{ReStyle3D\xspace}
\newcommand{\img}{\mathbf{I}\xspace}
\newcommand{\latent}{\mathbf{z}\xspace}
\newcommand{\dataset}{\textit{SceneTransfer}\xspace}

\newcolumntype{R}{>{\raggedleft\arraybackslash}X}

\hypersetup{
    colorlinks=true,
    urlcolor=red         % Change this to your preferred color
}

%% file: figures/tex/teaser.tex
\begin{teaserfigure}
 \includegraphics[width=\textwidth]{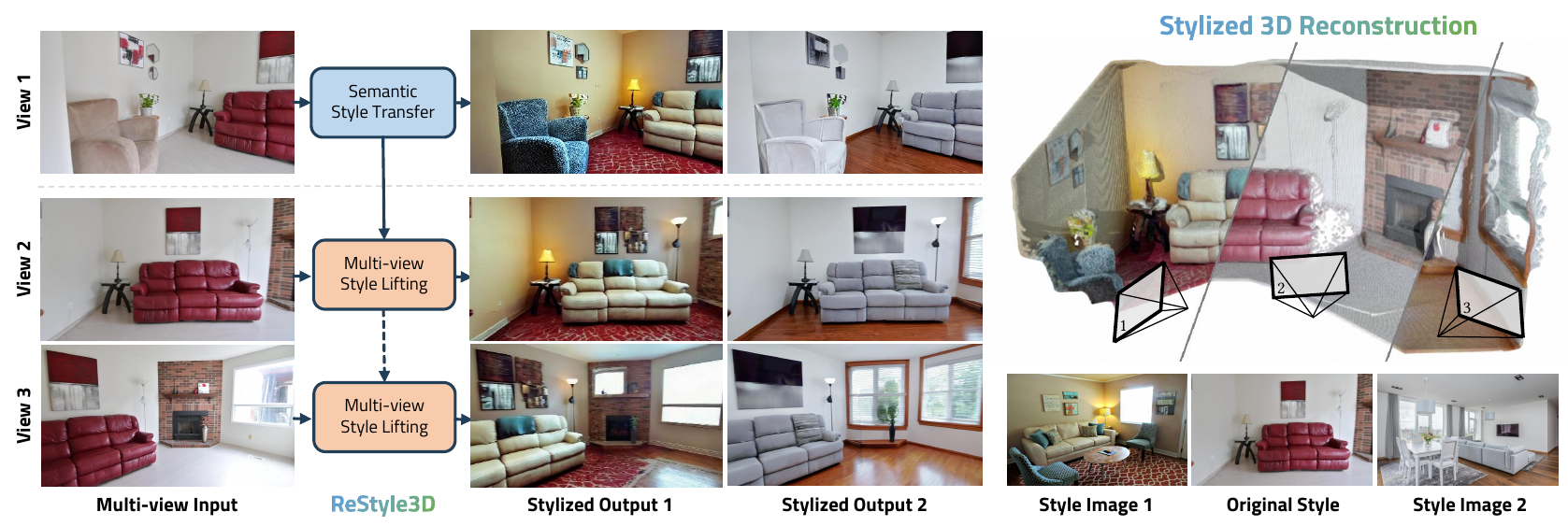}
  \caption{\textbf{\ours Overview}. Given an interior design image (style image) and a 3D scene captured by video or multi-view images, \ours first transfers the appearance based on semantic correspondences to a single view, then lifts the stylization to multiple viewpoints using 3D-aware style lifting, achieving multi-view consistent appearance transfer with fine-grained details.}
\label{fig:teaser}
\vspace{1em}
\end{teaserfigure}

%% file: sections/1_abstract.tex
\begin{abstract}
We introduce \ours, a novel framework for scene-level appearance transfer from a single style image to a real-world scene represented by multiple views. The method combines explicit semantic correspondences with multi-view consistency to achieve precise and coherent stylization.
Unlike conventional stylization methods that apply a reference style globally, \ours uses open-vocabulary segmentation to establish dense, instance-level correspondences between the style and real-world images. This ensures that each object is stylized with semantically matched textures.
\ours first transfers the style to a single view using a training-free semantic-attention mechanism in a diffusion model.
It then lifts the stylization to additional views via a learned warp-and-refine network guided by monocular depth and pixel-wise correspondences.
Experiments show that \ours consistently outperforms prior methods in structure preservation, perceptual style similarity, and multi-view coherence.
User studies further validate its ability to produce photo-realistic, semantically faithful results.
Our code, pretrained models, and dataset will be publicly released, to support new applications in interior design, virtual staging, and 3D-consistent stylization.  {\renewcommand\UrlFont{\color{magenta}\rmfamily\itshape} Project page and code at \url{https://restyle3d.github.io/}.}
\end{abstract}

%% file: sections/2_introduction.tex
\section{Introduction}
Generative diffusion models have recently spurred significant advances in image stylization and broader generative applications, enabling the seamless synthesis or editing of images with remarkable visual fidelity.
While existing image stylization approaches~\cite{chung2024styleid,li2024diffstyler} often excel at transferring well-known artistic styles~(\eg, Van Gogh paintings) onto photographs, they fall short when it comes to practical and realistic style applications, such as virtual staging or professional interior decoration, where transferring the style of one image (\textit{style} image) to another (\textit{source} image) entails transferring the individual appearance of objects (see Fig. \ref{fig:teaser}).

These methods tend to treat the style image globally, ignoring the semantic correspondence between individual objects or regions in the images.
This coarsely aligned stylization not only misrepresents object appearances but also fails to adapt fine-grained textures to semantically matched regions~(\eg, transferring \emph{couch} textures only to \emph{couches}). 
This is crucial for real-world use cases where style is defined by the unique characteristics (\eg, color, material, shape) of design elements (\ie, furniture, decor, lighting, and accessories) that give it its signature look \cite{park2022analysis}. 
Another line of work pursues \textit{semantic correspondence} for transferring object appearances~\cite{cheng2024zeroshot,zhang2023tale}.
While these methods show promise in aligning single objects or small regions via deep feature matching, they typically operate at low spatial resolutions~(often $64\times64$) and therefore struggle to handle complex scenes with strong perspective and multiple object instances.
Extending them to scene-level stylization remains a challenging problem due to both semantic and geometric complexity.

Moreover, when a scene is represented by multiple images (\eg, for larger coverage), ensuring multi-view \emph{consistency} in scene-level appearance transfer further complicates the task.
Existing multi-view editing methods~\cite{or24mvedit,styleGaussians,fujiwara2024sn2n,vicanerf2023} commonly require known camera poses and an existing 3D scene representation (\eg, a neural radiance field~\cite{mildenhall2021nerf} or 3D Gaussian splatting~\cite{kerbl20233d}), which needs a dense set of input views and considerable compute time. 
These methods struggle with sparse or casually captured views, and their specialized 3D pipelines hinder plug-and-play use.
A pixel-space approach preserving geometric cues without heavy 3D modeling is preferable but remains under explored. 
We propose \textbf{\ours}, a novel framework for scene-level appearance transfer that combines semantic correspondence and multi-view consistency, addressing limitations of 2D stylization and 3D-based editing methods.
\emph{Our key insight} is that the inherent but implicit semantic correspondences from pretrained diffusion models or vision transformers~(\eg, StableDiffusion~\cite{rombach2022latentdiffusion} and DINO~\cite{caron2021dino, oquab2023dinov2}) are insufficient for fine-grained, scene-level appearance transfer, especially when different objects or viewpoints are involved.
We tackle this by explicitly matching open-vocabulary panoptic segmentation predictions between the style and source images, while ensuring that unmatched parts of the scene still receive a global style harmonization.
This open-vocabulary labeling~(with no predefined semantic categories) helps us robustly align semantically corresponding regions even in cluttered indoor scenes.
By integrating these explicit correspondences into the attention mechanism of a diffusion process, we achieve more accurate and flexible stylization of multi-object scenes.

To further ensure \emph{3D awareness} and view-to-view consistency, we adopt a two-stage pipeline.
First, we achieve \emph{training-free} semantic appearance transfer in a single view by injecting our correspondence-informed attention into a pretrained diffusion model.
Second, a warp-and-refine diffusion network that efficiently propagates the stylized appearance to additional views in an auto-regressive manner, guided by monocular depth and pixel-level optical flows.
Our method does not require explicit pose or 3D modeling, and we show that the final stylized frames are fully compatible with off-the-shelf 3D reconstruction tools, enabling complete 3D visualizations and consistent multi-view stylization with minimal overhead.

In summary, our contributions are as follows:
\begin{itemize}
    \item We introduce \emph{SceneTransfer}, a new task of compositionally transferring multi-object appearance from a single style image to a 3D scene captured in multi-view images or video.
    \item We propose \ours, a two-stage pipeline that (\emph{i}) repurposes a pretrained diffusion model with \emph{semantic attention} for instance-level stylization, and (\emph{ii}) trains a warp-and-refine novel-view synthesis module to propagate the style across all views, maintaining global consistency.
    \item We create the \dataset benchmark with 25 interior design images and 31 indoor scenes~(243 style-scene pairs) from different categories (\eg bedroom, living room, and kitchen). Our results show strong improvements in structure preservation, style fidelity, and cross-view coherence.
\end{itemize}

%% file: sections/3_related_work.tex
\section{Related Work}
% Stylization has been a long-standing research topic, aiming to transfer artistic styles from a source to images while preserving the structural content.
% Our work intersects image stylization, semantic correspondence, and 3D-consistent novel-view synthesis.
% Classical stylization usually lacks object-level alignment, while semantic correspondence focuses on single objects or limited resolutions.
% Novel-view synthesis often relies on explicit 3D representations and dense camera poses.
% By leveraging open-vocabulary semantic matching and diffusion, we achieve scene-level appearance transfer with multi-view consistency, all without the overhead of 3D reconstruction.

\paragraph{Image Stylization} The goal is to transfer artistic styles to images while preserving structural content.
Early CNN-based methods~\cite{nst2016gatys, huang2017adain, dumoulin2017adversarially} laid the groundwork by capturing style and content representations.
With the advent of diffusion models~\cite{ho2020ddpm, rombach2022latentdiffusion}, recent approaches leverage pretrained architectures and textual guidance for high-quality stylization~\cite{chung2024styleid, subrtova2023diffusionimageanalogies, li2023stylediffusion, everaert2023diffusioninstyle, yang2023zero, li2024diffstyler, zhang2023inst}.
InST~\cite{zhang2023inst} employs textual inversion to encode styles in dedicated text embeddings, achieving flexible transfer.
StyleDiffusion~\cite{li2023stylediffusion} further refines style-content separation through a CLIP-based disentanglement loss applied during fine-tuning.
StyleID~\cite{chung2024styleid} adapts self-attention in pretrained diffusion models to incorporate artistic styles without additional training.
While these methods produce compelling results, they focus on overall style transfer without explicitly modeling semantic correspondences.
In contrast, we attempt to inject semantic matching in stylization, thereby enabling precise style transfer according to semantically matching regions.

\paragraph{Semantic Correspondence.}
Foundational works and recent innovations have shaped the evolution of semantic correspondence.
SIFT-Flow~\cite{liu2011siftflow} pioneered dense image alignment with handcrafted SIFT descriptors~\cite{lowe2004sift}.
Self-supervised vision transformers like DINO~\cite{caron2021dino} and DINO-V2~\cite{oquab2023dinov2, darcet2023vitneedreg} improved feature representation for semantic matching without labeled data~\cite{tumanyan2023disentangling,tumanyan2022splicing}.
Recent methods, such as ~\cite{zhang2023tale,hedlin2023keypoints}, DIFT~\cite{tang2023dift}, cross-image-attention~\cite{alaluf2024cross}, and~\cite{go2024eyeforaneye}, integrate diffusion models with these transformers, achieving superior zero-shot correspondence.
Techniques like Deep Functional Maps~\cite{cheng2024zeroshot} further refine correspondences by enforcing global structural consistency, demonstrating the potential of advanced representations in addressing correspondence challenges.
The development of these techniques enables the extraction of semantic correspondences using intermediate representations. 

\paragraph{Attention-based Control in Diffusion Models.}
The attention modules in pretrained diffusion models are essential in controlling the generated content, allowing various image editing tasks through attention mask manipulation.
Prompt-to-Prompt~\cite{hertz2022p2p} pioneered text-based local editing by manipulating cross-attention between text prompts and image regions.
Similarly, Plug-and-play~\cite{tumanyan2023plugandplay} leverages the original image's spatial features and self-attention maps to preserve spatial layout while generating text-guided edited images.
Epstein et al.~\shortcite{epstein2023diffusionselfguidance} introduced Diffusion Self-Guidance, a zero-shot approach that leverages internal representations for fine-grained control over object attributes.
While these methods focus on text-to-image attention control, recent works like Generative Rendering~\cite{cai2023genren} explore cross-image attention by injecting 4D correspondences from meshes into attention for stylized video generation. MasaCtrl~\cite{cao_2023_masactrl} proposed text-based non-rigid image synthesis by injecting attention masks between text and image.
In contrast, we propose a direct image-to-image semantic attention mechanism that transfers appearances across all semantic categories simultaneously through explicit correspondence masks, enabling efficient and accurate scene-level stylization without text prompts or 3D priors.

\paragraph{Diffusion-based Novel-View Synthesis (NVS)}
NVS of general scenes typically requires inferring and synthesizing new regions that are either unobserved or occluded in the original viewpoint.
A common strategy in prior work~\cite{wiles2020synsin, rockwell2021pixelsynth, liu2021infinitenature, koh2022simple} is to follow a warp-and-refine approach: estimate a depth map from the input image, warp the image to the desired viewpoint, and then fill in occluded or missing areas through a learned refinement stage.
More recent research~\cite{rombach2021geometryfree, yu2023photoconsistentnvs, jin2024lvsm} avoids explicit depth-based warping by directly training generative models that handle view synthesis in a single feed-forward pass. StoryDiffusion~\cite{zhou2024storydiffusion} proposes consistent self-attention to boost long-term consistency.
Another line of work~\cite{hao2023text2immersion, chung2023luciddreamer, shriram2024realmdreamer, cai2022diffdreamer, tseng2023poseguideddiffusion, yu2024viewcrafter, sun2024dimensionx, deng2024streetscapes, seo2024genwarp} integrates diffusion models such as StableDiffusion~\cite{rombach2022latentdiffusion}, making it possible to extrapolate plausible new views that are far from the input image for in-the-wild contents.
ReconX~\cite{liu2024reconx} and ViewCrafter~\cite{yu2024viewcrafter} both harness powerful video diffusion models combined with coarse 3D structure guidance to mitigate sparse-view ambiguities, achieving improved 3D consistency for novel-view synthesis.
Motivated by recent success in the warp-and-refine paradigm~\cite{seo2024genwarp}, we adopt a similar strategy but with a focus on style lifting, incorporating historical frames through adaptive blending to consistently propagate our style transfers across multiple views.

%% file: sections/4_method.tex
\section{ReStyle3D}
We present \ours, a framework for fine-grained \textit{appearance transfer} from a style image $\img_{style}\in \mathbb{R}^{H\times W \times3}$, to a 3D scene captured by \textit{unposed} multi-view images or video $\mathcal{X}_{src}:=\{\img^i_{src} \in \mathbb{R}^{H\times W \times3}\}_{i=1}^N$. Specifically, \ours aims to transfer the appearance of each region in $\img_{style}$ to its semantically corresponding region in $\mathcal{X}_{src}$, while maintaining multi-view consistency across all images. We assume spatial overlap between two consecutive frames in $\mathcal{X}_{src}$.

\subsection{Preliminaries}
% \textbf{Diffusion models} 
\paragraph{Diffusion models} Diffusion processes progressively add noise to an image $\img_0$ sampled from a data distribution $p_{\mathrm{data}}(\img)$, transforming it into Gaussian noise $\img_T$ over $T$ steps, following a variance schedule $\{\alpha_t\}_{t=1}^T$:
\begin{equation}
p(~\img_t~|~\img_0~) = \mathcal{N}(~\img_t;~\sqrt{\alpha_t}~\img_0, 1-\alpha_t\mathbf{I}~),
\end{equation}
where $\img_t$ represents the noisy image at timestep $t$. 
The \textit{reverse} process is performed by a denoising model $\epsilon_\theta(\cdot)$ that gradually removes noise from $\img_t$ to obtain cleaner $\img_{t-1}$. Here $\theta$ is the learnable parameters of the denoising model. During training, the denoising model is trained to remove noise following the objective function~\shortcite{ho2020ddpm}:
\begin{equation}
\mathcal{L} = \mathbb{E}_{\img_0,t \sim \mathcal{U}(T),\epsilon \sim \mathcal{N}(0, I)}||\hat{\epsilon}_{\theta} - \epsilon||_2^2,
\label{eq:2}
\end{equation}
where $\hat{\epsilon}_{\theta}=\hat{\epsilon}_\theta(\img_t, t, c)$, and $c$ is an optional input condition such as text, image mask, or depth information.
At inference stage, a clean image $\img:=\img_0$ is reconstructed
from a randomly sampled Gaussian noise $\img_T\sim \mathcal{N}(0, I)$ through an iterative noise-removal process. The cornerstone of modern image-based diffusion models is the latent diffusion model~\cite{rombach2022latentdiffusion} (LDM), where the diffusion process is brought to the latent space~\cite{vqgan_2021_CVPR} of a variational autoencoder (VAE). This approach is significantly more efficient compared to working directly in the pixel space. 
% By a slight abuse of notation, we use the same symbols for latent images.

\input{figures/tex/method_2dtransfer}
% \noindent\textbf{Attention layers}
\paragraph{Attention layers} Attention layers are fundamental building blocks in LDM. Given an intermediate feature map $F \in \mathbb{R}^{L \times d_h}$, where $L$ denotes the feature length and $d_h$ represents the feature dimension, the attention layer captures the interactions between all pairs of features through query-key-value operations:
\begin{equation}
\begin{split}
\phi &= \text{softmax}\left(\frac{Q'\cdot K'^T}{\sqrt{d_h}}\right)\cdot V' \\
Q' &= Q \cdot W_q, \quad K' = K \cdot W_k, \quad V' = V \cdot W_v,
\end{split}
\end{equation}
where $\phi$ is the updated feature map, $Q', K'$, and $V'$ are linearly projected representations of the inputs via $W_q$, $W_k$, and $W_v$, respectively. In self-attention, the key, query, and value originate from the same feature map, enabling context exchange within the same domain. For cross attention, the key and value come from a different source, facilitating information exchange across domains. In \ours, we tailor the self-attention layers specifically for semantic appearance transfer, while leaving the cross-attention layers unchanged.

\subsection{Appearance Transfer via Semantic Matching}
\label{sec:attention_appearance_transfer}
To transfer the appearance of $\img_{style}$ to $\img_{src}$, prior attempts also employing diffusion models~\cite{alaluf2024cross,zhang2023tale,cheng2024zeroshot} have primarily focused on single objects, and struggle with scene-level transfer involving multiple instances. Our key observation is that the implicit semantic correspondences in foundation models~\cite{rombach2022latentdiffusion,oquab2023dinov2} are insufficient for more complex multi-instance semantic matching. To address this limitation, \ours explicitly establishes and leverages semantic correspondences throughout the transfer process.

% \noindent\textbf{Open-vocabulary Semantic Matching.}
\paragraph{Open-vocabulary Semantic Matching.}
We leverage the open vocabulary panoptic segmentation model ODISE~\cite{xu2023open} for semantic matching. For a given input image, ODISE generates segmentation maps $\mathcal{M} \in \{1, \ldots, C\}^{H \times W}$, assigning each pixel to one of $C$ semantic categories. These maps enable semantic correspondences between the style and source images (detailed below). By matching open-vocabulary semantic predictions, \ours is not limited by predefined semantic categories in a scene. The correspondences are injected into the diffusion process to guide appearance transfer between matched regions.

\paragraph{Injecting Correspondences in Self-attention.}
\ours enables training-free style transfer by extending the self-attention layer of a pretrained diffusion model (Fig.~\ref{fig:method_2d}). This approach injects style information from $\img_{style}$ into $\img_{src}$ while preserving its structure. Specifically, we first encode both the style and source images into the latent space of Stable Diffusion~\cite{rombach2022latentdiffusion}, producing $\latent^{style}_0$ and $\latent^{src}_0$. These latent representations are then inverted to Gaussian noise, $\latent^{style}_T$ and $\latent^{src}_T$, using edit-friendly DDPM inversion~\cite{huberman2024edit}.
To enhance structural preservation and mitigate LDM’s over-saturation artifacts, we incorporate monocular depth estimates~\cite{yang2024depth} of the input images through a depth-conditioned ControlNet~\cite{zhang2023adding} during the inversion process. The stylized image latent is then initialized as $\latent^{out}_T = \latent^{src}_T$.

Next, we transfer the style from $\latent^{style}_T$ to $\latent^{out}_T$ by de-noising them along parallel paths~\cite{alaluf2024cross}. At each de-noising step $t$, we extract style features $(K_{style}, V_{style})$ and query features $Q_{out}$ from individual self-attention layers. The semantic-guided attention for the output feature $\phi_{out}$ is computed by combining the attention features with the attention mask $M$ as follows:
\begin{equation} 
\phi_{out} = \text{softmax}\left(\frac{Q_{out} \cdot K_{style}^T}{\sqrt{d_h}} \odot M \right) \cdot V_{style}, 
\end{equation}
where $\odot$ denotes element-wise multiplication and $\phi_{out} \in \mathbb{R}^{d^2 \times d_h} $ is passed to the next layer after self-attention.

To obtain the attention mask $M \in \mathbb{R}^{d^2 \times d^2}$, we flatten and bilinearly downsample the semantic masks $\mathcal{M}_{style}$ and $\mathcal{M}_{src}$ to match the resolution of attention feature maps, which is $d\times d$. The attention mask is defined as $M(i,j)=1$ if the $i$-th region in the source and the $j$-th region in the style image share the same semantic class; otherwise, $M(i,j)=0$. 
This formulation ensures that each region in the output image samples its appearance solely from semantically corresponding regions in the style image.
For example (Figure \ref{fig:attn_vis}), a rug in the source image is only cross-attended to its counterpart in the style image, inheriting its appearance.
If multiple instances in the style image share the same semantic class, attention is distributed across them based on sampling weights determined by softmax attention scores.
This mechanism naturally extends to support user-specified correspondences.
Regions without semantic matches attend to the entire style image to preserve global harmony.
While semantic attention effectively transfers appearance, it may compromise realism and structure, requiring further refinement.

\input{figures/tex/attn_vis}
\paragraph{Guidance and Refinement.}
We draw inspiration from~\cite{ho2021cfg,alaluf2024cross} and incorporate classifier-free guidance~(CFG) combined with semantic and depth-conditioned generation. At each denoising step $t$, we compute three noise predictions: $\epsilon_t$, $\epsilon_t^{d}$ and $\epsilon_t^{s}$. Here, $\epsilon_t^{s}$ represents the predicted noise from the semantic attention path, $\epsilon_t^{d}$ is obtained from the depth-conditioned ControlNet~\cite{zhang2023adding}, and $\epsilon_t$ is the unconditional noise prediction. The final noise prediction is then calculated as follows: 
\begin{equation}
\hat{\epsilon}_t = (1 - \alpha) \epsilon_t + \alpha (\lambda_s\epsilon_t^s + \lambda_d\epsilon_t^d),
\end{equation}
where $\lambda_s$ and $\lambda_d$ are the respective guidance weights ($\lambda_s+\lambda_d=1$) for semantic and depth guidance. $(1 - \alpha)$ is the classifier-free guidance scale, which balances conditional and unconditional predictions, improving image realism.

To enhance image quality, we employ a two-stage refinement process. First, we upscale the initial stylized image from 512×512 to 1024×1024 resolution. Then, following SDEdit~\cite{meng2022sdedit}, we add high-frequency noise to this upscaled image and denoise it for 100 steps with SDXL~\cite{podell2024sdxl}. This refinement process enhances local details while maintaining the overall style, producing our final single-view output $\hat{\img}_{src}$.
% \shengyu{I suggest to expand this part a bit, to make it more self-contained.}

\input{figures/tex/independent_stylization}
\input{figures/tex/method_lifting}

\subsection{Multi-view Consistent Appearance Transfer}\label{sec:multiview_style_propagation}
While our semantic attention module effectively transfers the appearance for a single view, applying this independently to each view may cause inconsistent artifacts~(see Fig~\ref{fig:inconsistency}). Thereby, we develop an approach to transfer the appearance of the stylized image $\hat{\img}_{src}^i$ to all remaining views, while maintaining multi-view consistency. 

\paragraph{Flow-guided Style Warping.}
Given a pair of source images $(\img_{src}^i, \img_{src}^j)$, we first leverage the stereo matching method~\cite{wang2024dust3r} to extract the dense point correspondence and the camera intrinsics. Using these, the optical flow $\mathbf{W}_{i\rightarrow j}\in \mathbb{R}^{H \times W \times 2}$ is calculated by projecting the pointmaps of $i$-th image to the $j$-th image. Next, given the optical flow and the stylized $i$-th image $\hat{\img}_{src}^i$, we employ softmax splatting~\cite{softslatting_CVPR_2020} to obtain the initial stylized image $\hat{\img}_{w}^j$ and its warping mask $\mathbf{M}_{w}^j$, which indicates missing pixels in the $j$-th frame after forward warping. 

\input{tables/comparison_2d}
\paragraph{Learning View-to-View Style Transfer.}
Given the source image $\img_{src}^j$ and its initial stylized version $\hat{\img}_{w}^j$, we train a 2-view warp-and-refine model $\hat{\epsilon}_{\theta}=\hat{\epsilon}_\theta(\latent_t, t, c)$ to generate a complete and consistent stylized image following conditions $c$: the initial stylized image, the inpainting mask, and the monocular depth map $\mathbf{D}^j$ of the source image $\img_{src}^j$ (Fig.~\ref{fig:method_lifting}). The final condition $c=\mathrm{concat}(\latent^{(\hat{\img}_{w}^j)}, \latent^{(\mathbf{M}_{w}^j)}, \latent^{(\mathbf{D}^j)})$, $\latent^*$ denotes individual latent representations. To harness the power of a pretrained diffusion model~\cite{podell2024sdxl}, like~\cite{ke2023marigold}, we modify the input channels of its initial convolution layer to accommodate additional conditions and zero-initializing the additional weights. Following Eq.~\eqref{eq:2}, we train the model using quadruplets of the warped and incomplete image, depth map, mask, and the clean and complete image. The model learns to complete missing pixels, while globally refining all pixels to address warping artifacts.

% \noindent\textbf{Auto-regressive Multi-view Stylization}
\paragraph{Auto-regressive Multi-view Stylization.}
We propose an autoregressive approach to extend two-view stylization to handle multiple views or even videos, ensuring global coherence across the scene (Fig.~\ref{fig:method_lifting}). Stylizing the $j$-th frame using only the previous frame $(j-1)$ can lead to inconsistencies with earlier frames while warping all historical frames could produce blurry outputs. Instead, we warp the stylized frame $(j-1)$ along with two randomly selected historical frames. In overlapping regions, where multiple pixels are warped to the same location, we adopt an exponential weighted averaging to blend pixels, prioritizing pixels from frame $(j-1)$. This adaptive weighting ensures temporal consistency while preserving sharp details in the resulting warped image $\hat{\img}_{w}^{1:j-1}$. Finally, our model refines the output, producing a fully stylized frame.

%% file: figures/tex/method_2dtransfer.tex
\begin{figure}[t!]
\centering
\includegraphics[width=\linewidth]{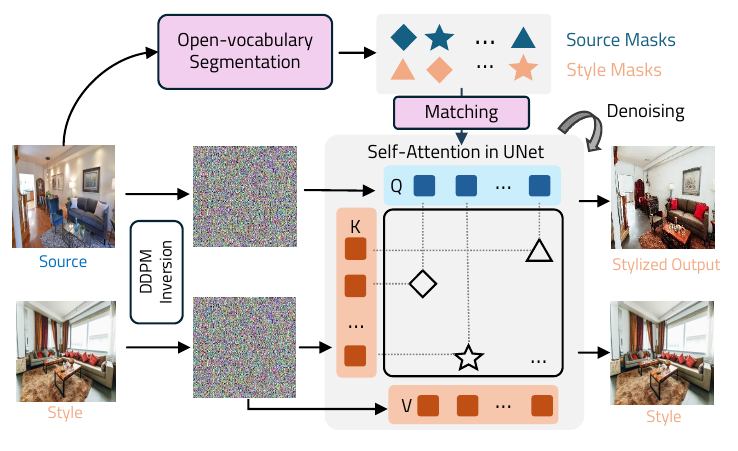}
% \caption{\textbf{2D Semantic Appearance Transfer}. The style and source images are inverted back to step $T$ using DDPM~\shortcite{huberman2024edit}. We extract the semantic correspondence mask between the two images by open-vocabulary semantic segmentation and label matching. This mask then guides the information exchange between the source image and style image.
% } 
\caption{\textbf{Semantic Appearance Transfer}. 
The style and source images are first noised back to step $T$ using DDPM inversion~\shortcite{huberman2024edit}. During the generation of the stylized output, the extended self-attention layer transfers style information from the style to the output latent. This process is further guided by a semantic matching mask, which allows for precise control.
} 
\label{fig:method_2d}
\end{figure}

%% file: figures/tex/attn_vis.tex
\begin{figure}[h]
    \centering
        \includegraphics[width=0.99\linewidth]{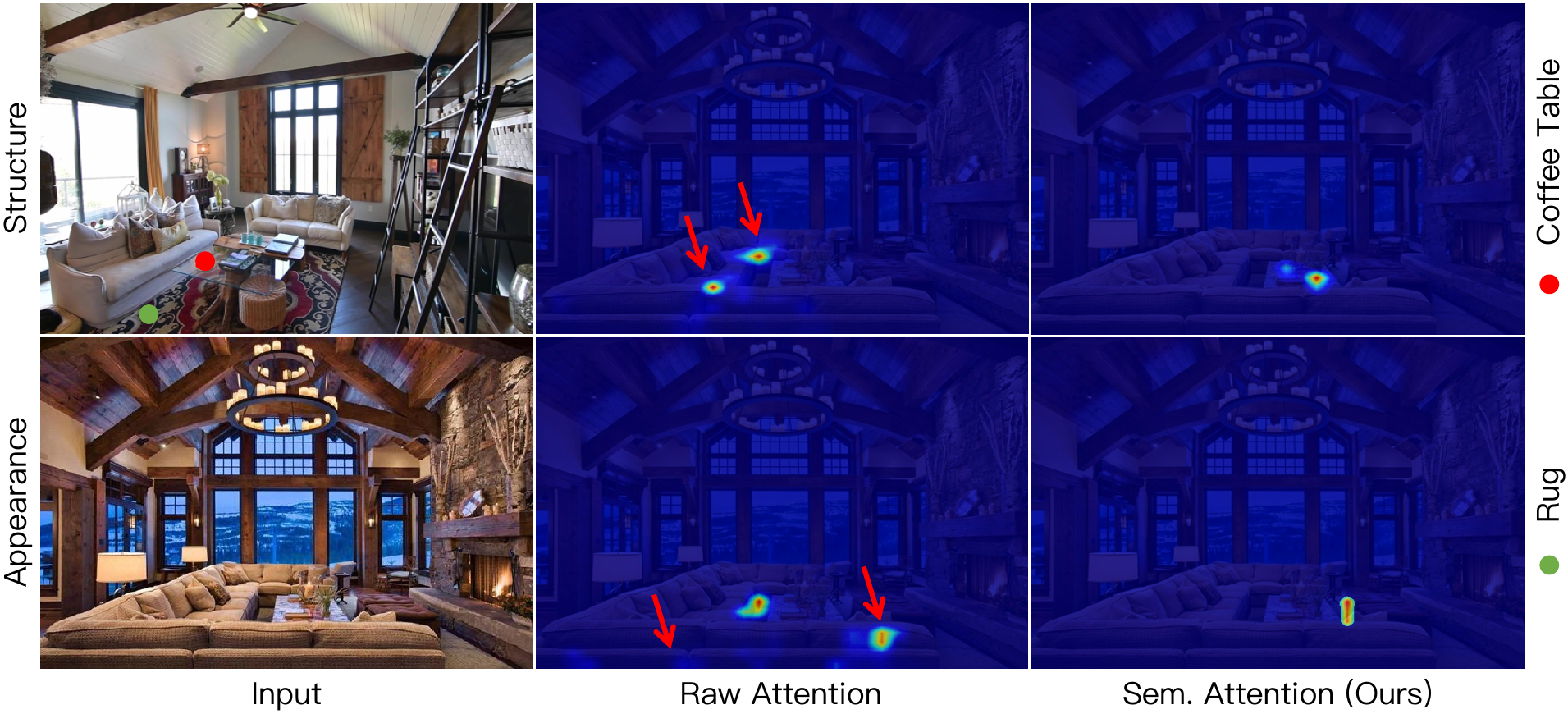}
\caption{\textbf{Attention Query Visualization}. We visualize the attention score at two query positions, coffee table and rug. Raw attention in \cite{alaluf2024cross} spilled across regions (red arrows) due to multi-instance ambiguity, semantic attention effectively confines the activation in the matched region.} 
\label{fig:attn_vis}
% \vspace{-2mm}
\end{figure}

%% file: figures/tex/independent_stylization.tex
\begin{figure}[t]
    \centering

        \includegraphics[width=\linewidth]{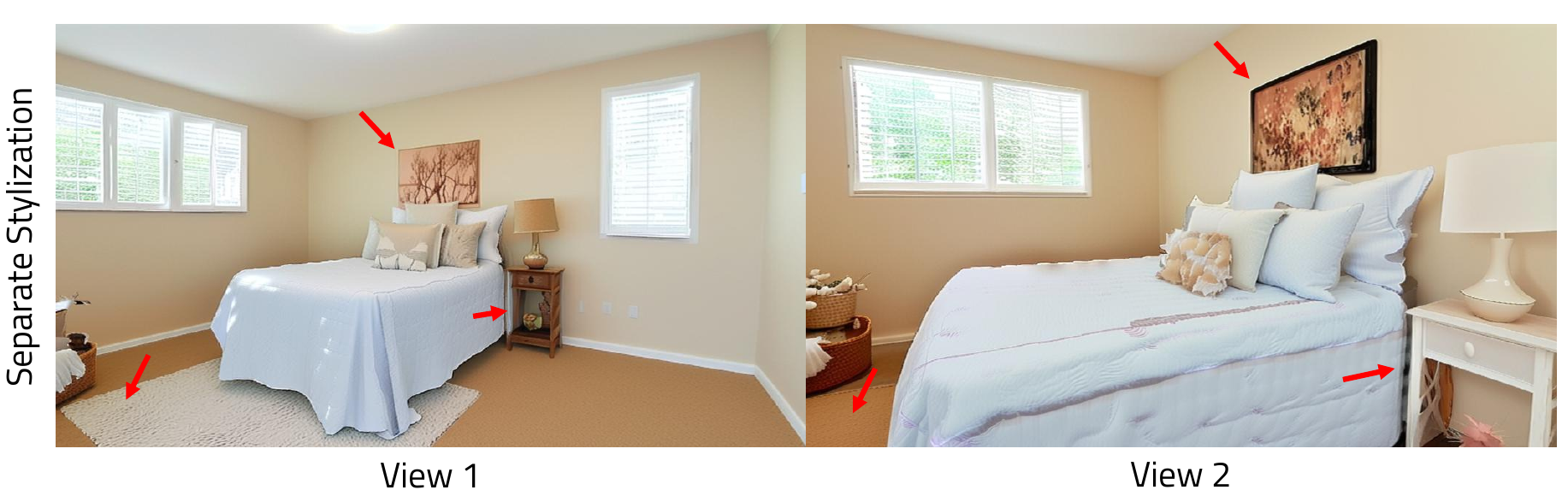}
\caption{\textbf{Multi-view Inconsistency Caused by Separate Transfer}. When stylizing each view separately, we observe inconsistencies in the results (highlighted by red arrows) due to high variance in generative modeling.} 
\label{fig:inconsistency}
\end{figure}

%% file: figures/tex/method_lifting.tex
\begin{figure}[t!]
\centering
\includegraphics[width=\linewidth]{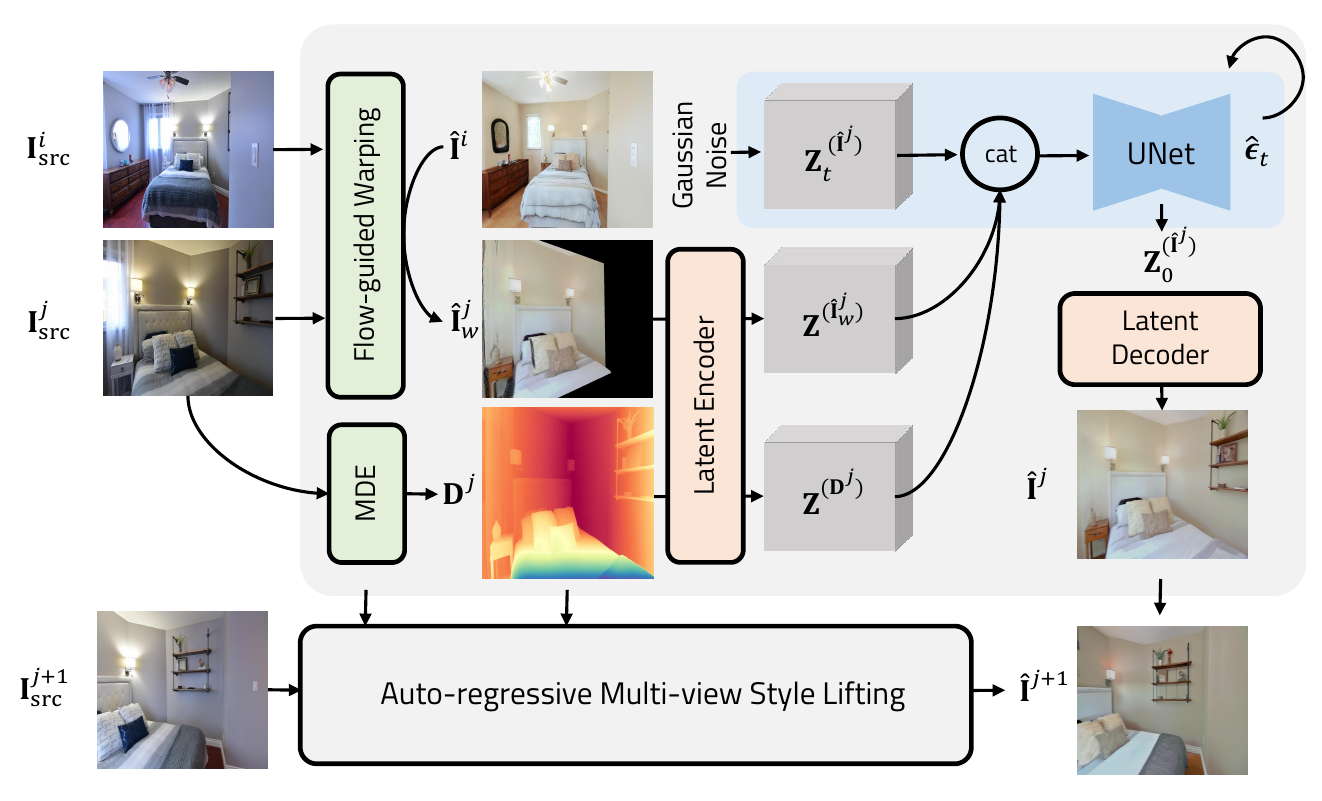}
% \caption{\textbf{Multi-view Style Lifting}. Stereo correspondences are extracted from the original image pair $(\img_{src}^i, \img_{src}^j)$ and used to warp the stylized image $\hat{\img}_{src}^i$ to the second image ($\img_{src}^j$). We train a warp-and-refine model to complete the style on the second image. This model is then applied iteratively to multiple views in an auto-regressive manner.}
\caption{\textbf{Multi-view Style Lifting}. Stereo correspondences are extracted from the original image pair $(\img_{src}^i, \img_{src}^j)$ and used to warp the stylized image $\hat{\img}^i$ to the second image, $\img^j_w$. To address missing pixels from warping, we train a warp-and-refine model to complete the stylized image $\hat{\img}^j$. This model is applied across multiple views within our auto-regressive framework.}
\label{fig:method_lifting}
\end{figure}

%% file: tables/comparison_2d.tex
\begin{table*}[t]
\caption{
\textbf{Quantitative Comparison of \ours and Baseline Methods on 2D Appearance Transfer}. Our method achieves the best overall performance for both structure preservation and perceptual similarity, benefiting from its explicit semantic guidance and two-stage refinement.
}
\centering
\resizebox{0.96\linewidth}{!}{
\setlength{\tabcolsep}{12pt}
\begin{tabular}{l|ccc|ccc|c}
\toprule
\multirow{2}{*}{\textbf{Method}}& \multicolumn{3}{c|}{Depth Metrics (\textbf{Structure})}& \multicolumn{3}{c|}{Perceptual Similarity (\textbf{Style})}& \multirow{2}{*}{\textbf{Avg. Rank}}\\
\cmidrule{2-7}
&AbsRel$\downarrow$&SqRel$\downarrow$&$\delta_1$$\uparrow$&DINO$\uparrow$&CLIP$\uparrow$&DreamSim$\downarrow$&\\
\midrule
Cross-Image-Attn.~\cite{alaluf2024cross} & 22.47 & 7.944 & 5.78 & 0.553 & 0.709 & 0.414 & 4.8 \\
IP-Adatper SDXL~\cite{ye2023ip} & 9.38 & 1.847 & 79.29 & 0.570 & 0.752 & 0.371 & 3.0 \\
StyleID~\cite{chung2024styleid} & 11.25 & 2.59 & \textbf{93.44} & 0.546 & 0.741 & 0.332 & 3.2 \\
\midrule
\ours (Ours w/o refinement) & 11.30 & 2.65 & 89.11 & \textbf{0.586} & 0.778 & 0.319 & 2.5 \\
\ours (Ours w/ refinement) & \textbf{8.34} & \textbf{1.67} & 88.45 & 0.584 & \textbf{0.783} & \textbf{0.316} & \textbf{1.5} \\
\bottomrule
\end{tabular}
}
\label{tab:2d_transfer}
\end{table*}

%% file: sections/5_experiments.tex
\section{Experiments}

\input{figures/tex/2d_transfer}
\paragraph{Implementation Details.}
We base our semantic attention module on Stable Diffusion 1.5~\cite{rombach2022latentdiffusion} and the refinement and 2-view warp-and-refine model on SDXL~\cite{podell2024sdxl}. To train our two-view warp-and-refine model~(Sec.~\ref{sec:multiview_style_propagation}), we use 4 NVIDIA A100 40GB GPUs with an effective batch size of 256 for 20K iterations, using the AdamW optimizer~\cite{loshchilov2018adamw} with learning rate $10^{-4}$. We randomly drop out half of the text prompt during training to make our model agnostic to text conditions. The model is trained on a dataset with 57K house tour images featuring 57 different houses and apartments.

\subsection{Evaluation Setting}
\paragraph{Dataset.} Our \dataset benchmark comprises 31 distinct indoor scenes captured as short video clips, totaling 15,778 frames across multiple room categories, including living rooms, kitchens, and bedrooms, all disjoint from our training data. To evaluate stylization capabilities, we curated a set of 25 interior design reference images, enabling 243 unique style-scene combinations. Evaluation is performed on 1,109 keyframes sampled from these clips. For more details on data, please refer to the supplementary material (Supp.).

\paragraph{Evaluation Metrics.}
We evaluate multiple different aspects of our pipeline.
First, we assess the appearance transfer performance using source images on two aspects: structure preservation and style transfer quality.
For structure preservation, we compare depth maps predicted by DepthAnythingV2~\cite{yang2024depth} between stylized and original images using standard metrics: Absolute Relative Error~(AbsRel), $\delta1$ accuracy, and Squared Relative Error~(SqRel), following established protocols~\cite{ke2023marigold, yang2024depth}.
For style transfer quality, we measure perceptual similarity between the stylized output and the style image using DINOv2~\cite{oquab2023dinov2}, CLIP, and DreamSim~\cite{fu2023dreamsim} scores. We evaluate this task on the stylized source images of each scene.
Next, we evaluate our two-view lifting model~(Sec.~\ref{sec:multiview_style_propagation}).
We assess its warp-and-refine quality using PSNR, SSIM~\cite{zhou2004ssim}, and LPIPS~\cite{zhang2018lpips} while also reporting FID~\cite{martin2017fid} to quantify the realism of generated frames under challenging viewpoint extrapolation. We evaluate using pairs of the source images per scene and their warped projections on the rest of the frames in each scene---we exclude pairs without correspondences. We do not use any stylization to train or evaluate since there is no ground truth.
To evaluate global consistency, we leverage DUSt3R~\cite{wang2024dust3r} to extract poses by aligning point maps from stylized sequences and compute cumulative error curve~(AUC) by comparing recovered camera poses against those from original images.

\subsection{Results}
\paragraph{Image Appearance Transfer.}
We compare with three state-of-the-art methods on image-conditioned stylization and appearance transfer: Cross Image Attention~\cite{alaluf2024cross}, IP-Adapter~\cite{ye2023ip}, and StyleID~\cite{chung2024styleid}. For a fair comparison, we add depth ControlNet~\cite{zhang2023adding} to SDXL IP-Adapter~\cite{ye2023ip} and use the style image as the image prompt. As shown in Tab.~\ref{tab:2d_transfer}, our method achieves superior performance on both structure preservation and style transfer metrics. Notably, our explicit semantic attention mechanism in the diffusion UNet enhances the perceptual similarity between stylized outputs and style images, as evidenced by better DINO, CLIP, DreamSim scores, and attention visualization~(Fig.~\ref{fig:attn_vis}). The refinement step further improves structure preservation, reducing AbsRel from 11.30 to 8.34 and SqRel from 2.65 to 1.67.
Qualitative comparisons (Figs.~\ref{fig:2d_transfer} and~\ref{fig:2d_transfer_more}) reveal the limitations of existing approaches. Cross Image Attention effectively captures style textures but fails to maintain scene structure due to the lack of semantic guidance. IP-Adapter SDXL preserves overall structure but struggles with local detail transfer, as it compresses style information into a global feature vector. While StyleID achieves the second-best performance, its results tend to keep high-frequency details from the source image while applying style changes more globally, showing limited capability in fine-grained appearance transfer.

We conduct a user study with 27 participants who were shown examples of a source and style image with outputs from four methods.
Participants selected the result that best preserved the structure while faithfully transferring the style.
Out of 252 evaluations (Tab. \ref{tab:user_study}), \ours was the most preferred, achieving the highest preference (42.4\%) and demonstrating its effectiveness in balancing structure preservation with appearance transfer under human perception.

\input{tables/user_study}
\input{tables/pairwise_nvs}
\input{figures/tex/multiview_transfer}

\paragraph{Two-view NVS.}
We compare our approach to: \textit{i)}~SDXL inpainting model~\cite{podell2024sdxl} with depth-conditioned ControlNet~\cite{zhang2023adding}, \textit{ii)} GenWarp~\cite{seo2024genwarp}, an image-based diffusion model for single view NVS, \textit{iii)} StoryDiffusion~\cite{zhou2024storydiffusion}, a model with consistent self-attention for long-range image and video generation and \textit{iv)} ViewCrafter~\cite{yu2024viewcrafter}, a video-diffusion model for NVS. Note that the proposed task differs from traditional NVS as it leverages geometry information from the \textit{novel view} itself. We employ DUSt3R~\cite{wang2024dust3r} to extract the correspondences and provide the initial warped image as input to all methods. As shown in Tab.~\ref{tab:2viewnvs}, \ours outperforms  across all metrics, achieving a superior reconstruction ability as evidenced by the best PSNR, SSIM, and LPIPS metrics. Additionally, it exhibits strong capability in extending style to unseen regions, evidenced by the lowest FID score (Fig.~\ref{fig:scannet_recon}). Notably, the second best method ViewCrafter~\cite{yu2024viewcrafter}, requires a predefined camera trajectory as input to video diffusion and runs 10$\times$ slower than ours. 

\input{tables/pose_eval}
\paragraph{Multi-view Consistency Evaluation.}
We further evaluate the multi-view consistency of the stylized results through a proxy task Specifically, we input the original and stylized images to DUSt3R~\shortcite{wang2024dust3r} and estimate the camera poses, separately. By evaluating the agreement with the poses from the original images, we analyze whether the geometry is preserved in the stylized images. As shown in Tab.~\ref{tab:corres_select}, our adaptive auto-regressive approach effectively mitigates inconsistencies while preserving image sharpness, significantly outperforming the baselines on all pose metrics. 
Figs.~\ref{fig:scannet_recon} and~\ref{fig:recon_extra} show multi-view results, including the 3D reconstruction of stylized outputs with estimated camera poses, demonstrating both geometric and style consistency despite camera motion and multiple objects. StoryDiffusion~\cite{zhou2024storydiffusion} does not support multi-view stylization.

\paragraph{Ablation Study.}
In Tab. \ref{tab:ablation}(a), we run \ours without our guidance strategy and observe significant degradation in structure preservation (AbsRel from 8.34 to 16.72). In (b), removing semantic attention hurts performance on perceptual similarity \wrt style image, showing that both components are crucial for semantic-accurate style transfer while maintaining structural integrity. Attending unmatched instance to the style image globally provides better style fidelity compared to keeping to original image, noted as Keep. Attn..

\input{tables/ablation}

%% file: figures/tex/2d_transfer.tex
\begin{figure*}[t]
    \centering
 \begin{tabular}{cccccc}
    \hspace{0.6cm} \small \textbf{Source Image} & \hspace{1.0cm} \small \textbf{Style Image} & \hspace{0.8cm}\small \ours (Ours) & \hspace{.6cm}\small Cross-Image-Attn. & \hspace{0.7cm}\small IP-Adapter SDXL & \hspace{0.3cm}\small StyleID \\
    \multicolumn{6}{c}{%
        \includegraphics[width=\linewidth]{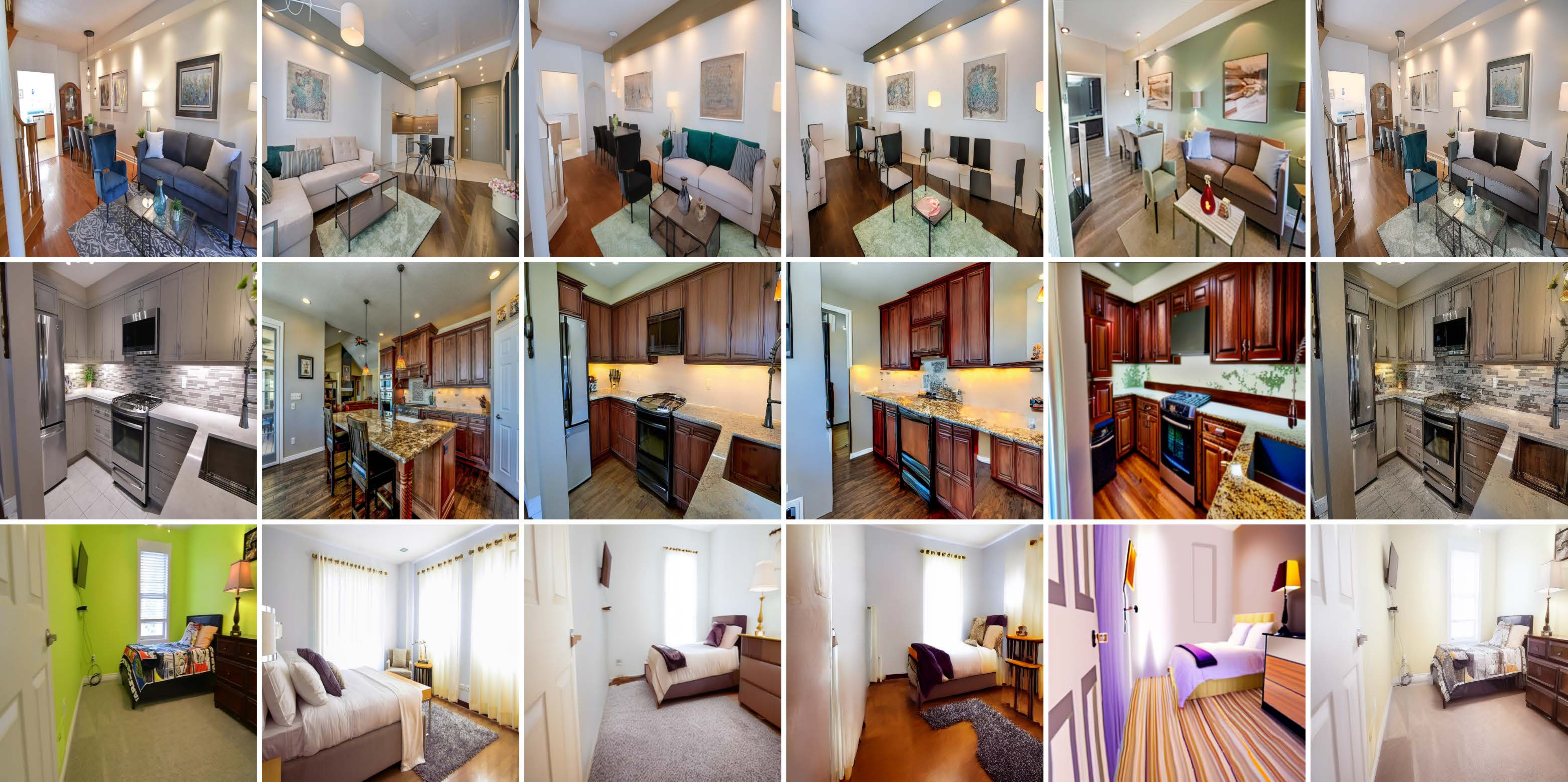}%
    } \\

\end{tabular}  
   \caption{\textbf{Image Appearance Transfer Results}. Our method enables precise appearance transfer between semantically corresponding elements, evidenced by the green rug and glass table (first row), textured cabinet (second row), and bedsheets (third row). Unlike baselines that either apply global style transfer or fail to preserve structure, \ours maintains both semantic fidelity and structural integrity.}
    \label{fig:2d_transfer}
\end{figure*}

%% file: tables/user_study.tex
\begin{table}[h]
\caption{\textbf{Image Appearance Transfer User Study.} We show user preference rates (\%) for different methods, where participants selected the result that best preserved the original scene structure while closely matching the reference style. \ours achieves the highest preference rate.}
\label{tab:user_study}
\resizebox{\linewidth}{!}{
\begin{tabular}{lcccc}

\toprule
\textbf{Method} & ReStyle3D (Ours) & Cross Image Attn. & IP-Adapter & StyleID \\
%\midrule
\textbf{Preferred Rate} (\%) & \textbf{42.4} & 16.3 & 4.4 & 36.9 \\
\bottomrule
\end{tabular}}
% \vspace{-2mm}
\end{table}

%% file: tables/pairwise_nvs.tex
\begin{table}[t]
\caption{\textbf{Results on Two-view Novel-view Synthesis.} \ours achieves the highest scores on all metrics, indicating more accurate view synthesis and visually pleasing outputs compared to existing methods.}
\centering
\resizebox{\linewidth}{!}{
\begin{tabular}{l|c|cccc}
\toprule
\textbf{Method} & Res. & PSNR$\uparrow$ & SSIM$\uparrow$ & LPIPS$\downarrow$ & FID$\downarrow$ \\
\midrule
GenWarp~\cite{seo2024genwarp} & \multirow{5}{*}{512$^2$} & 13.503 & 0.465 & 0.435 & 59.965 \\
StoryDiffusion~\cite{zhou2024storydiffusion} & & 14.023 & 0.481 & 0.502 & 203.83 \\
SDXL Inpainting~\cite{podell2024sdxl} & & 16.228 & 0.535 & 0.389 & 89.502 \\
ViewCrafter~\cite{yu2024viewcrafter} & &  17.178 & 0.594 & 0.278 & 56.127 \\
\textbf{\ours (Ours)} & & \textbf{18.614} & \textbf{0.677} & \textbf{0.246} & \textbf{34.138} \\
\midrule
GenWarp~\cite{seo2024genwarp} & \multirow{5}{*}{1024$^2$} & 13.491 & 0.565 & 0.440 & 60.540 \\
StoryDiffusion~\cite{zhou2024storydiffusion} & & 14.014 & 0.583 & 0.476 & 198.32 \\
SDXL Inpainting~\cite{podell2024sdxl} & & 16.153 & 0.565 & 0.426 & 89.537 \\
ViewCrafter~\cite{yu2024viewcrafter} & & 17.137 & 0.652 & 0.317 & 57.898\\
\textbf{\ours (Ours)} & & \textbf{18.568} & \textbf{0.711} & \textbf{0.283} & \textbf{35.721} \\
\bottomrule
\end{tabular}}
\label{tab:2viewnvs}
% \vspace{-3mm}
\end{table}

%% file: figures/tex/multiview_transfer.tex
\begin{figure*}[t!]
    \centering
 \begin{tabular}{ccccc}
    % Image spanning all columns
      & \hspace{3.6cm} \small Frame 1 & \hspace{3.0cm}\small Frame 2 & \hspace{2.8cm}\small Frame 3 & \hspace{1cm}\small 3D Reconstruction  \\
    
    \multicolumn{5}{c}{%
        \includegraphics[width=\linewidth]{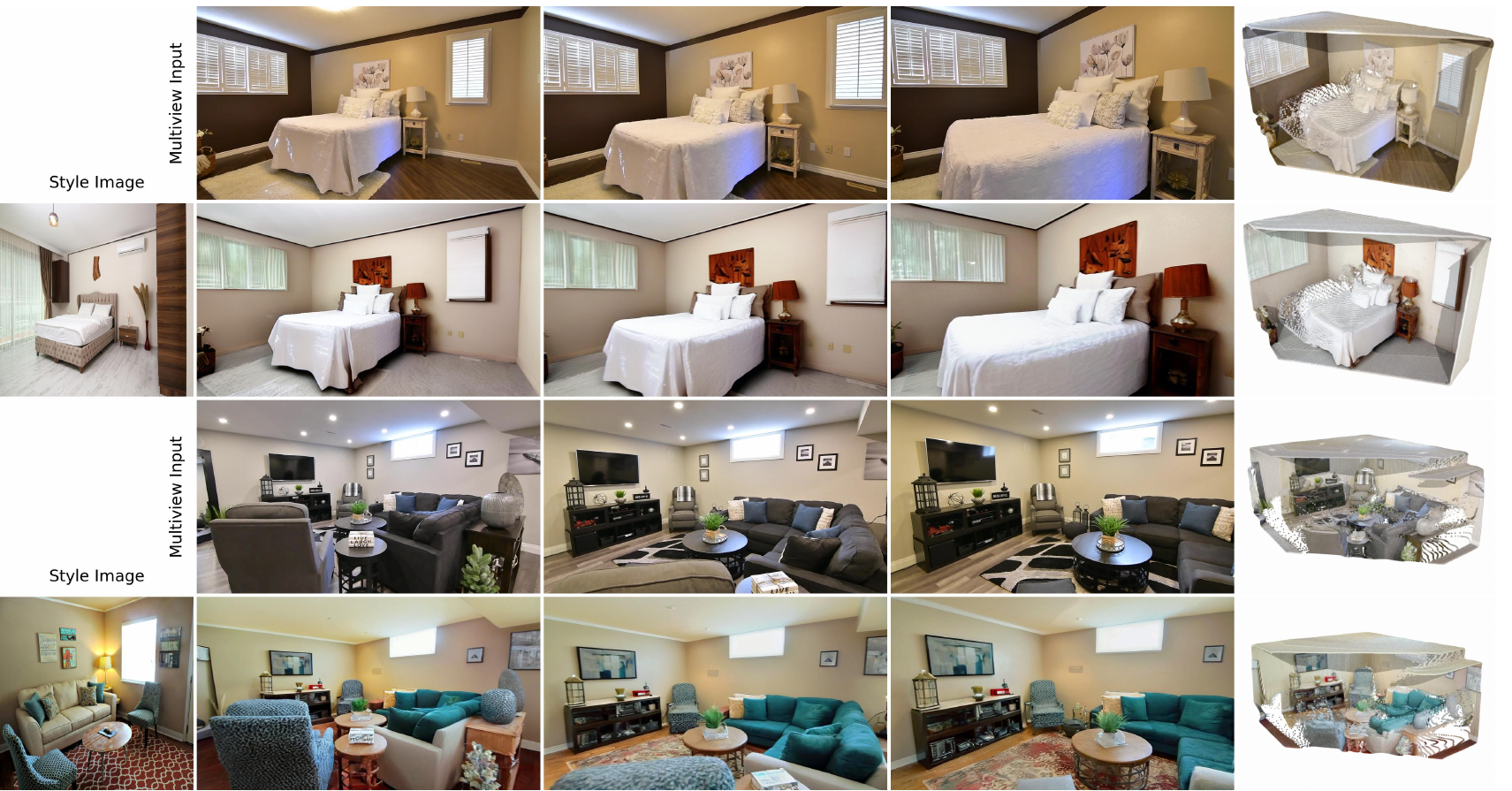}%
    } \\
    
\end{tabular}  

    \caption{\textbf{Results on Video/Multi-view Appearance Transfer of \ours}. We show the style images, three frames stylized by \ours, followed by a 3D reconstruction of these outputs using an off-the-shelf pipeline. Despite challenging camera motion and multiple objects in the scene, our method preserves consistent geometry and seamlessly transfers the reference style across all frames.} 
    \label{fig:recon}
\label{fig:scannet_recon}
% \vspace{1mm}
\end{figure*}

%% file: tables/pose_eval.tex
\begin{table}[t]
\caption{\textbf{Pose Deviation from Real-world Estimates.} We measure the fraction of camera poses within certain rotation~(at $5^\circ$, $10^\circ$, $15^\circ$) and translation~(at $1\mathrm{cm}$, $2\mathrm{cm}$, $5\mathrm{cm}$) error thresholds, reporting area-under-curve~(AUC) values. \ours achieves significantly higher AUC in both, showing superior multi-view geometric consistency vs. existing methods. }
\centering
\resizebox{\linewidth}{!}{
\begin{tabular}{l|ccc|ccc}
\toprule
\multirow{2}{*}{\textbf{Method}}& \multicolumn{3}{c|}{Rotation AUC$\uparrow$} & \multicolumn{3}{c}{Translation AUC$\uparrow$} \\
\cmidrule{2-7}
  & @5$^\circ$ & @10$^\circ$ & @15$^\circ$ & @1cm  & @2cm & @5cm\\
\midrule
GenWarp~\cite{seo2024genwarp}  & 25.89 & 46.70 & 58.89 & 58.38 & 59.39 & 70.05 \\
SDXL Inpainting~\cite{podell2024sdxl}  & 34.52 & 52.79 & 66.50 & 61.42 & 65.99 & 74.11 \\
ViewCrafter~\cite{yu2024viewcrafter}  & 37.56 & 55.33 & 68.53 & 60.91 & 65.99 & 77.16 \\
\textbf{\ours (Ours)} & \textbf{52.79} & \textbf{69.54} & \textbf{79.70} & \textbf{66.50} & \textbf{77.66} & \textbf{83.25} \\

\bottomrule
\end{tabular}}
\label{tab:corres_select}
\vspace{-5pt}
\end{table}

%% file: tables/ablation.tex
\begin{table}[t]
\centering
\caption{\textbf{Ablation Study.} We separately remove the guidance strategy and the semantic attention module to evaluate their impact on both structure preservation and style fidelity. Removing either significantly degrades performance, highlighting the importance of both components in achieving robust scene geometry and perceptually faithful stylization.}
\vspace{3pt}
\begin{minipage}{.7\linewidth}
\centering
\resizebox{\linewidth}{!}{
\begin{tabular}{lccc}
\toprule
& AbsRel$\downarrow$ & SqRel$\downarrow$ & $\delta_1 \uparrow$ \\
\midrule
Ours w/o guidance & 16.72 & 4.36 & 67.46 \\
Ours w/ guidance & \textbf{8.34} & \textbf{1.67} & \textbf{88.45} \\
\bottomrule
\end{tabular}
}
\vspace{1mm}
\subcaption{Ablation on Guidance}
\end{minipage}%

\begin{minipage}{.8\linewidth}
\centering
\resizebox{\linewidth}{!}{
\begin{tabular}{lccc}
\toprule
& DINO$\uparrow$ & CLIP$\uparrow$ & DreamSim$\downarrow$ \\
\midrule
Ours w/o Sem. Attn. & 0.492 & 0.682 & 0.419 \\
Ours w/ Keep. Attn. & 0.549 & 0.737 & 0.359 \\
Ours w/ Sem. Attn. & \textbf{0.584} & \textbf{0.783} & \textbf{0.316} \\
\bottomrule
\end{tabular}
}
\vspace{1mm}
\subcaption{Ablation on Semantic Attention}
\end{minipage}
\label{tab:ablation}
\end{table}

%% file: sections/6_conclusion.tex
\section{Conclusion}
We presented \ours, a framework for compositional semantic appearance transfer from a design image to multi-view scenes. 
Our two-stage approach combines training-free semantic attention in diffusion models with a multiview style propagation network to ensure semantic and geometric consistency across views. 
\ours avoids assumptions about scene semantics or geometry, making it suitable for real-world interior design and virtual staging.

\textbf{Limitations and Future Work. }\ours faces several challenges: \textit{i)} Drastic lighting changes between style and source images can confuse appearance transfer, \textit{ii)} Small objects are missed by the segmentation model. We discuss more in supplementary material.

%% file: figures/tex/refinement.tex
\begin{figure*}[t]
    \centering
 \begin{tabular}{cccc}
    \hspace{1.2cm} \small \textbf{Source Image} & \hspace{2.1cm} \small \textbf{Style Image} & \hspace{2.4cm}\small w/o refinement & \hspace{1.2cm}\small w/ refinement  \\
    \multicolumn{4}{c}{%
        \includegraphics[width=0.95\linewidth]{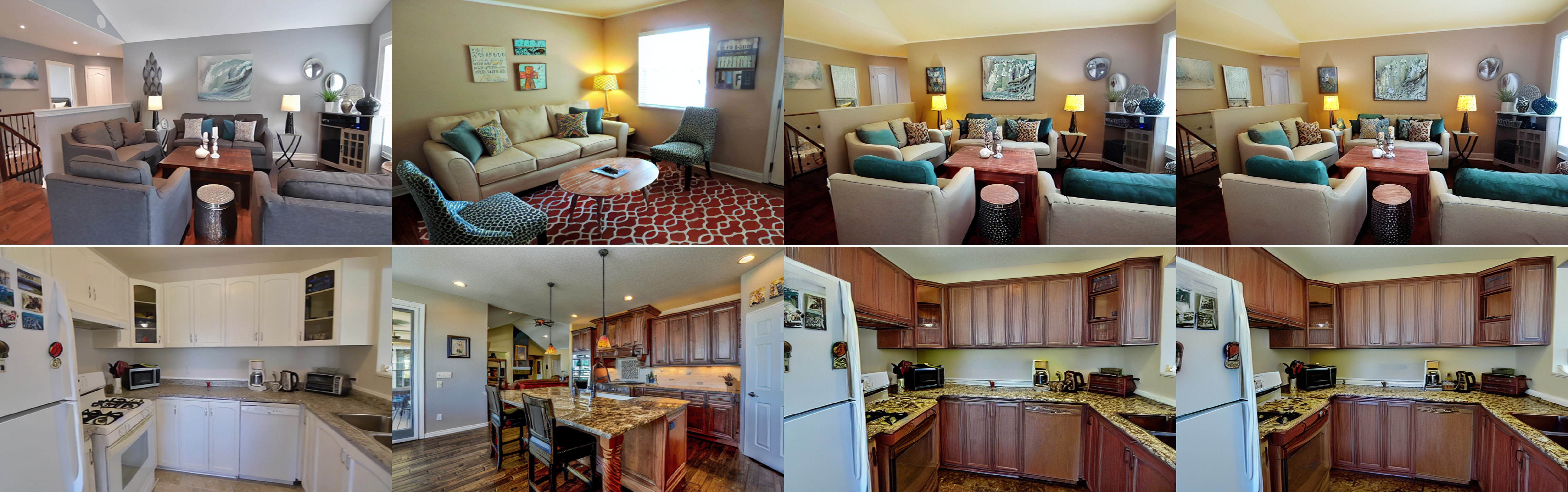}%
    } \\

\end{tabular}  
    \caption{\textbf{Qualitative Comparison on refinement module of \ours}. Results w/o refinement module present visually unpleasant artifacts on small objects and edge of the objects (\eg, candles on the coffee table in the first row, kettle and toaster in the second row.).  Our proposed refinement module (right) can effectively improve the quality in both color and geometry, while maintaining the global style consistent.} 
    \label{fig:2d_refinement}
\end{figure*}

%% file: figures/tex/2d_transfer_extra.tex
\begin{figure*}[t]
    \centering
 \begin{tabular}{cccccc}
    \hspace{0.6cm} \small \textbf{Source Image} & \hspace{1.0cm} \small \textbf{Style Image} & \hspace{0.8cm}\small \ours (Ours) & \hspace{.6cm}\small Cross Image Attn. & \hspace{0.7cm}\small IP-Adapter SDXL & \hspace{0.3cm}\small StyleID \\
    \multicolumn{6}{c}{%
        \includegraphics[width=\linewidth]{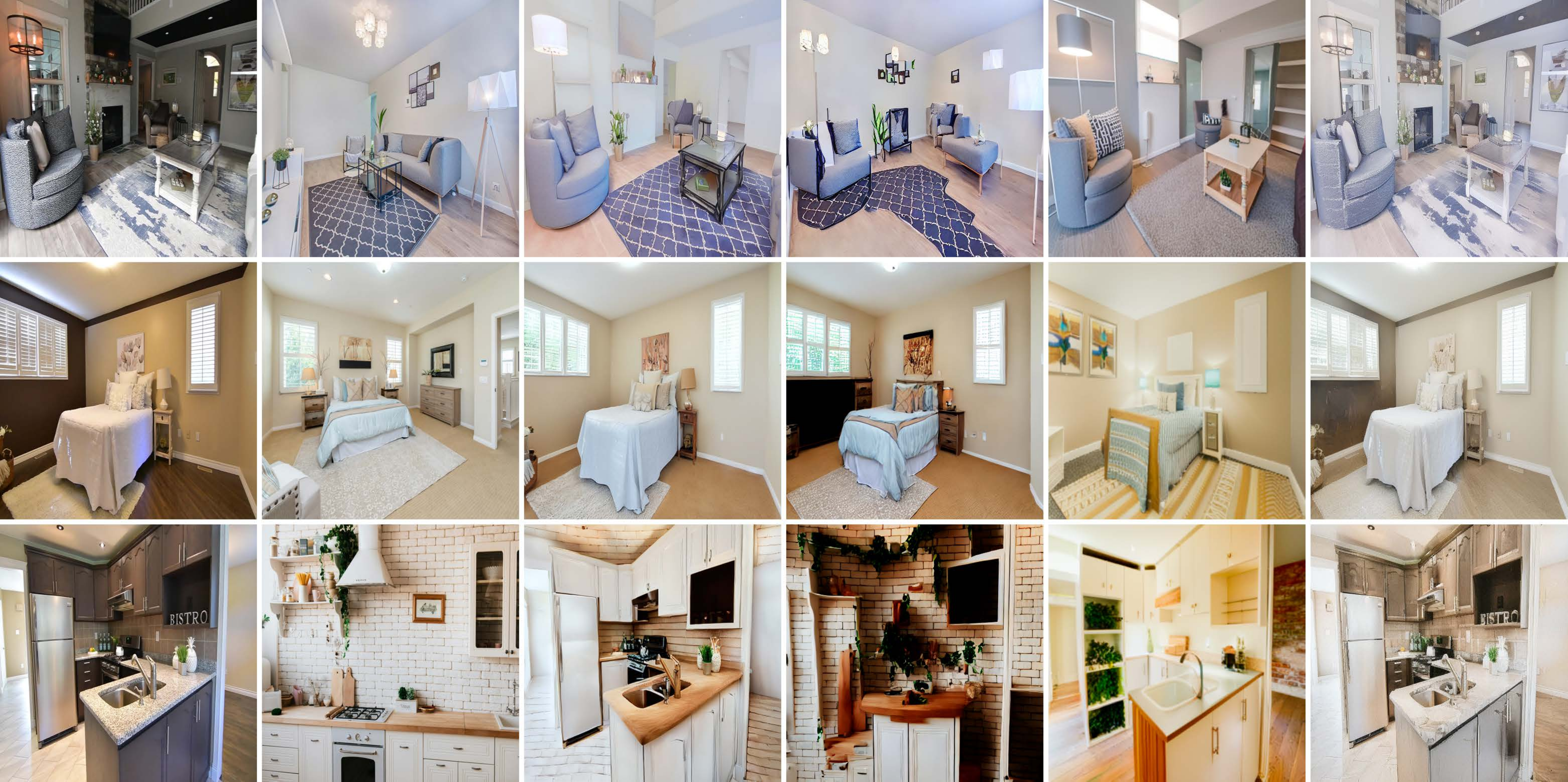}%
    } \\

\end{tabular}  
    \caption{\textbf{Additional results on 2D appearance transfer}. Each example shows the source image, the reference style image, and the stylized outputs. While the baseline methods either disrupt scene structure or misalign local style details, \ours consistently preserves geometric fidelity and correctly maps the reference appearance to each semantic region. Subtle details like furniture textures and decorative elements are accurately adapted to match the style.} 
    \label{fig:2d_transfer_more}
\end{figure*}

%% file: figures/tex/style_lifting.tex
\begin{figure*}[t!]
    \centering
 \begin{tabular}{cccccc}
    % Image spanning all columns
    \hspace{0.8cm} \small Input View & \hspace{1.3cm} \small GenWarp & \hspace{1.2cm}\small SDXL Inpainting & \hspace{1.cm}\small Viewcrafter & \hspace{1.cm}\small \ours (Ours) & \hspace{0.2cm}\small Reference View \\
    
    \multicolumn{6}{c}{%
        \includegraphics[width=\linewidth]{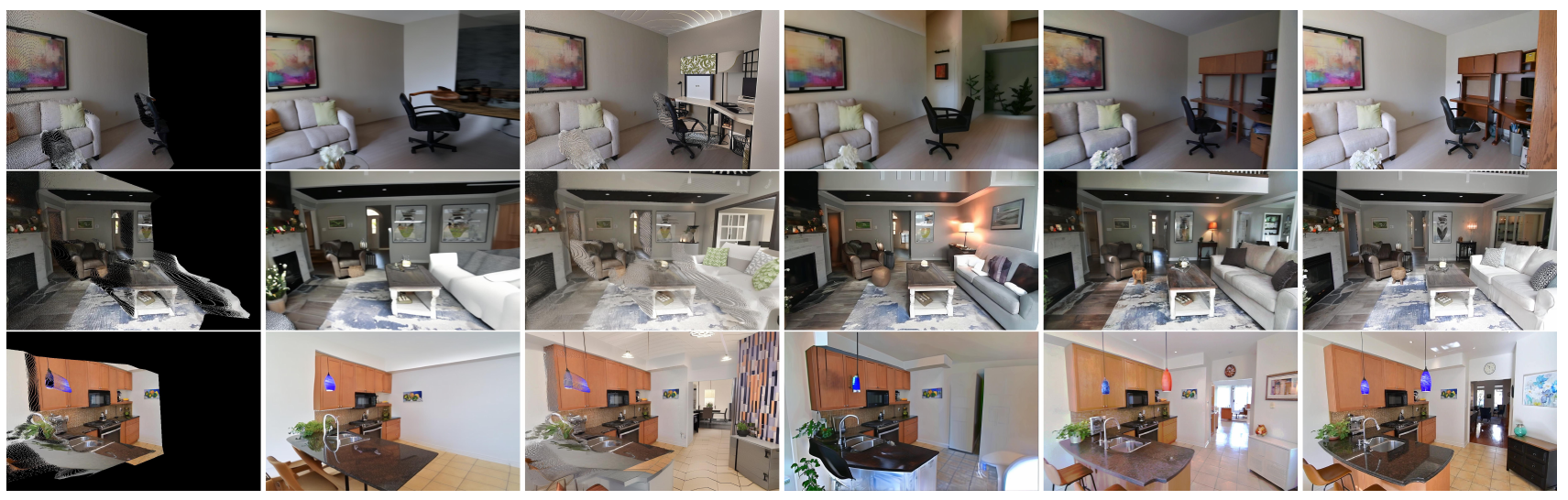}%
    } \\
    
\end{tabular}  

    \caption{\textbf{Results on two-view NVS with warp-and-refine}. Given a single input view and a target viewpoint, each method attempts to synthesize the target frame by warping and refining the source image. \ours recovers more accurate geometry and fewer artifacts, while also preserving finer scene details. By contrast, baseline methods struggle with consistent edge alignment and realism, showing noticeable artifacts and incomplete regions.}
    %\label{fig:recon}
\label{fig:scannet_recon}

\end{figure*}

%% file: figures/tex/extra_end2end.tex
\begin{figure*}[t!]
    \centering
 \begin{tabular}{ccccc}
    % Image spanning all columns
      & \hspace{3.6cm} \small Frame 1 & \hspace{3.0cm}\small Frame 2 & \hspace{2.8cm}\small Frame 3 & \hspace{1cm}\small 3D Reconstruction  \\
    
    \multicolumn{5}{c}{%
        \includegraphics[width=\linewidth]{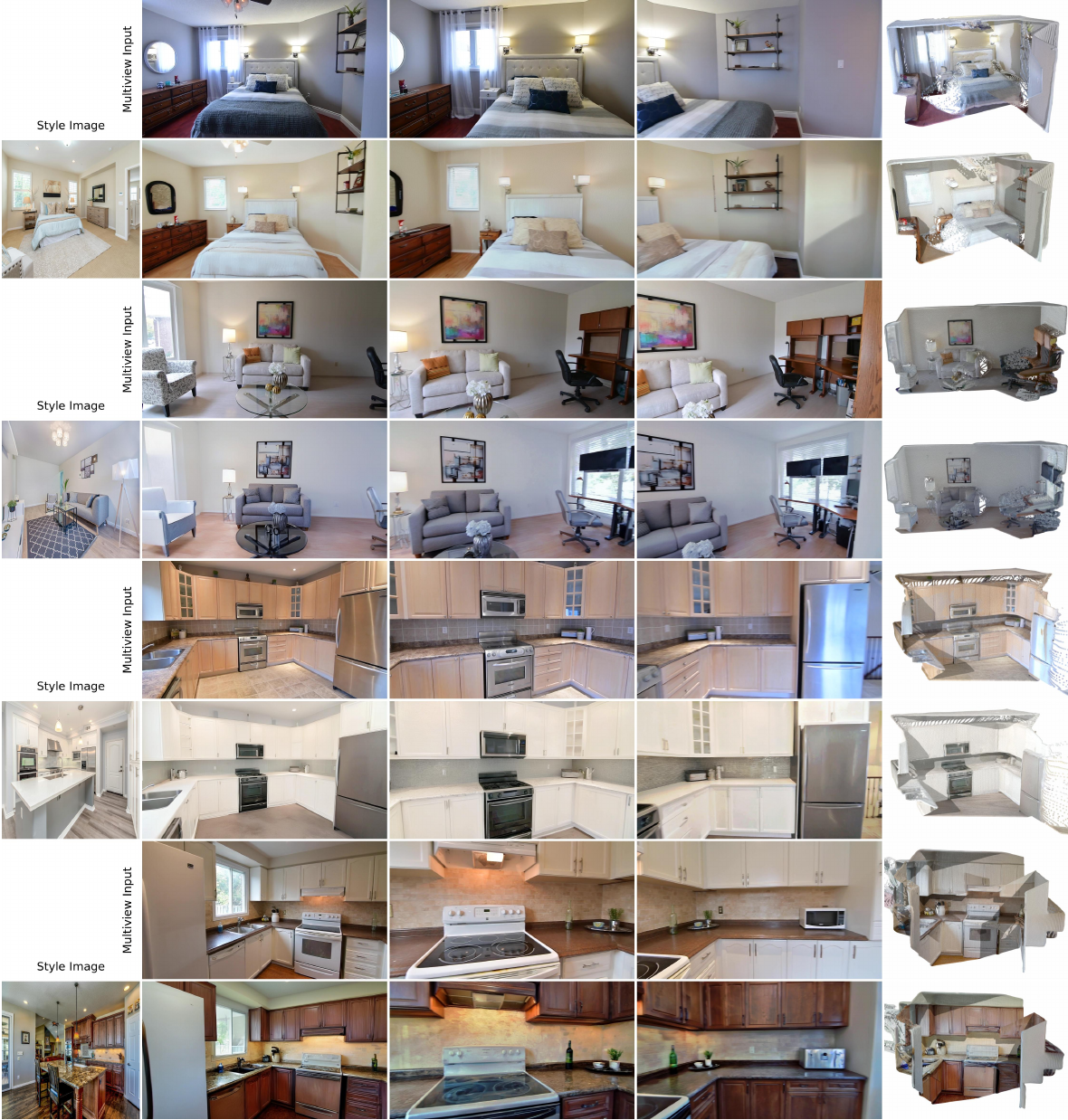}%
    } \\
    
\end{tabular}  

    \caption{\textbf{Additional Results on Video/Multiview Appearance Transfer}. We showcase three frames from a new indoor sequence stylized by \ours, followed by a 3D reconstruction of these stylized images using an off-the-shelf algorithm. Despite dynamic viewpoint changes and scene complexity, \ours consistently enforces semantic correspondences and preserves geometric integrity across all frames, enabling high-quality multi-view edits for practical applications such as interior design or virtual staging.} 
    \label{fig:recon_extra}

\end{figure*}